\documentclass{article}
\usepackage{iclr2027_conference,times}
\usepackage{amsmath,amssymb}
\usepackage{graphicx}
\usepackage{booktabs,tabularx}
\usepackage{float}
\usepackage{xcolor}
\usepackage{hyperref}
\usepackage{url}
\usepackage{enumitem}

\title{The COUNT: \raisebox{-0.15em}{\includegraphics[trim=214bp 167bp 214bp 220bp,clip,height=1.2em]{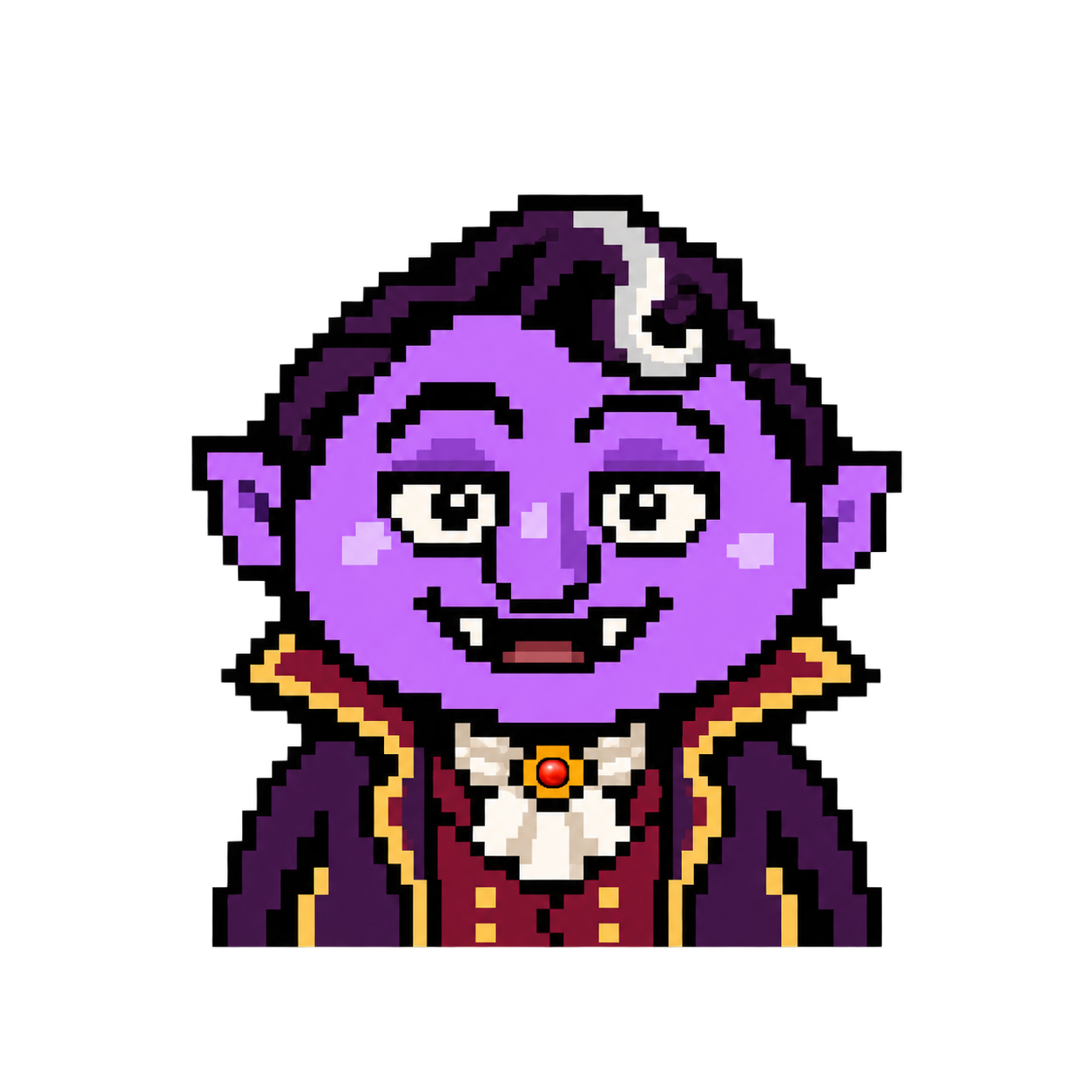}}\\[0.25em]  Counting N-grams Unlocks Scaling on Host Memory and Training-Free Adaptation}
\author{Ian Magnusson$^{1}$ \quad Noah A. Smith$^{1,2}$ \quad Pang Wei Koh$^{1}$ \\
$^{1}$University of Washington \quad $^{2}$Allen Institute for AI
}

\iclrfinalcopy
\begin{document}
\maketitle
\lhead{}

\begin{abstract}
The scaling relationship of compute and performance for language models (LMs) has been extensively studied; more recently, models are being designed to scale memory use, as well. In particular, adding embeddings for n-grams scales memory separately from compute use, because each pass only processes individual embedding rows, which are much smaller than the storage used for the whole embedding table.
However, existing methods train embeddings from scratch, which needs expensive GPU memory for fast reading and writing.
Meanwhile, the ordinary memory of the host system is cheap and underutilized by these training methods.
To scale on host memory, we introduce COUNT (Counting Offline Unlocks N-gram Transformers), which builds n-gram embeddings without adding learned parameters by storing precomputed n-gram counts on host. We use the corpus frequencies to weight together the LM's own unembeddings of the next tokens most often following the current n-gram history. At a fixed compute budget, COUNT can be added as an additional input embedding for pretraining 1B-parameter LMs to improve validation loss as much as an LM with 57\% more compute. If GPU memory is limited, using COUNT to let backbone parameters fill GPU memory gives better performance than splitting memory with additional learned n-gram embeddings.
Beyond efficiency, COUNT allows switching in a different table of n-gram counts from a domain-specific corpus at inference time to adapt to a new task without any training.
\end{abstract}

\section{Introduction}
\label{sec:introduction}

Compute and memory are both crucial resources for scaling language models (LMs). The tradeoff of compute and performance has been improved through optimal allocation of compute to parameters and training tokens~\citep{kaplan2020scaling,hoffmann2022training}. Beyond compute, performance can also be bought by increased \emph{memory} through architectures that scale sparsely activated parameters, such as by activating only a subset of experts in mixture-of-experts (MoE) models~\citep{shazeer2017outrageously,lepikhin2020gshard,fedus2022switch}.
The scaling of another type of sparsely activated parameters, input embeddings, has recently seen much exploration~\citep{roy2022ngrammer,huang2025over,liu2026scaling,cheng2026engram}. By embedding n-grams as well as individual tokens, many new parameters can be added to the embedding table while each lookup still processes only one row. Thus, compute scales only with the width of the embedding not the much larger total size of the embedding table.

Existing n-gram embedding work learns representations from scratch jointly with pretraining. Consequently, they are not easily offloaded from expensive GPU memory during training, because their weights and optimizer state must be read and written quickly, increasing communication complexity.
Therefore these learned n-gram embeddings compete for GPU memory with the backbone model's parameters.
But a long line of language modeling research and settled knowledge has built representations of n-grams without expensive pretraining by instead counting the frequency of n-grams in a corpus~\citep{jelinek1980interpolated,kneser1995improved,brants2007large}.
These methods refine the basic principle that the likelihood of a continuation can be estimated from counts of its occurrence after an n-gram history that is a suffix of the current context.
Existing work integrates such count-based language modeling or related corpus-retrieval methods with neural LMs~\citep{khandelwal2020knnlm, li2022residual,liu2024infinigram}. But unlike scaling dense input representations, these fuse raw next-token distributions at the output.

We propose \emph{COUNT}, Counting Offline Unlocks N-gram Transformers, which builds n-gram embeddings with precomputed corpus next-token statistics stored in host RAM.
COUNT connects the expressivity of learned embeddings with the read-only storage of count-based scores.
For each n-gram history, corpus counts supply a fixed list of frequent next tokens.
The backbone LM's unembedding head already learns representations of these predicted next tokens, so COUNT uses frequency scores to weight and combine unembeddings of next tokens as a new input embedding summed with the token embedding.
This adds no learned parameters, and since the large counts table is read-only, it does not need to use GPU memory. Since COUNT's count tables are computed offline, they can also be cheaply recomputed over new corpora to change its embeddings at inference time with no additional parameter training.

We train COUNT on a range of backbone LMs from 60M--1B parameters and token to parameter ratios from 10 to 40, allowing us to observe how COUNT interacts with traditional parameter and token scaling. We find that standard transformers without COUNT would need 46--89\% more compute to achieve the same language modeling loss as COUNT and 6--56\% more compute to match downstream task loss. Moreover, COUNT achieves Pareto-optimal tradeoffs between loss and learned parameter count in comparisons with learned n-gram embeddings, which need more GPU RAM to increase performance comparably. COUNT also improves performance more than output-level fusion of corpus counts. We also experiment with swapping in target-domain n-gram statistics at inference time and get training-free improvements in domain-specific evaluations in math, biomedicine, and law. We release\footnote{\url{https://github.com/IanMagnusson/COUNT}} n-gram counts of the full Dolma 3 pretraining corpus~\citep{olmo2025olmo3}, along with training code and pretrained checkpoints for COUNT and learned n-gram baselines.

\begin{figure}[!t]
 \centering
 \includegraphics[width=\linewidth]{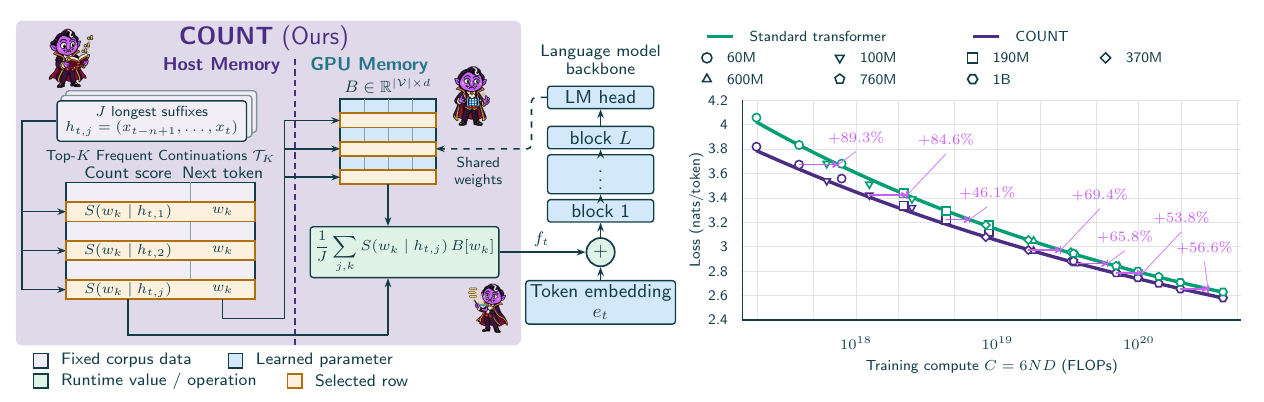}
 \caption{\textbf{Left:} COUNT uses corpus n-gram frequencies to score continuations and gather rows of the unembedding matrix into an additional input embedding, $f_t$. The count score $S(w\mid h)$ can be stupid backoff~\citep{brants2007large}, or any other frequency score based on $\operatorname{count}(h,w)/\operatorname{count}(h)$. Unlike n-gram embeddings learned from scratch, COUNT's additional storage can be offloaded to host RAM at training time. \textbf{Right:} COUNT reaches validation loss equivalent to a standard transformer trained with 46–89\% more compute. It does this without increasing compute by instead scaling n-gram statistics on otherwise unused host RAM by up to 797 GB. Solid curves are the compute-optimal fits of each method's scaling law. Each violet guide shows the extra compute required to make a standard transformer reach the loss of an optimally trained COUNT run. Figure~\ref{fig:six-arm-quality} shows task loss and downstream accuracy.}
 \label{fig:count-architecture}
\end{figure}

\section{Method}
\label{sec:method}

\subsection{Preliminaries}
\label{sec:preliminaries}

A language model predicts the next token $x_{t+1}$ from prefix $x_{1:t}$, where each token is an integer ID in a vocabulary $\mathcal V$.
Let $d$ be the model's hidden width (i.e., size of each vocabulary item's embedding vector), $E\in\mathbb R^{|\mathcal V|\times d}$ its token embedding matrix, and $e_t=E[x_t]$ the embedding of the token at position $t$.
We write $h_t^{(0)}\in\mathbb R^d$ for the corresponding input to the first transformer block; conventionally, $h_t^{(0)} = e_t$.

\subsection{COUNT}
\label{sec:ngram-fusion}

COUNT (Figure~\ref{fig:count-architecture}) constructs n-gram embeddings as an additional input to the LM backbone without adding any learned parameters. We use a precomputed, frozen table $\mathcal T_K$ that maps each of $H$ stored n-gram histories to its $K$ most frequent next tokens and their scores, derived cheaply from n-gram counts and approximating conditional probabilities of next tokens.
$K\ll|\mathcal V|$, the number of continuations retained per history, is chosen to retain just the head of the next-token distribution, where the n-gram LM is more robust.
Let $h_{t,j}$ be the longest stored suffix of $x_{1:t}$ at the $j$-th highest order of $n$.
The lookup $\mathcal T_K[h_{t,j}]$ returns its top-$K$ next tokens $w_k$ and their count scores $S(w_k\mid h_{t,j})$, such as the stupid-backoff scores in Equation~\ref{eq:backoff}.

We use the LM's unembedding head as a basis $B\in\mathbb R^{|\mathcal V|\times d}$, whose rows we weight by these scores to form the COUNT embedding $f_t$. Reusing this matrix avoids adding any learned parameters for COUNT. The continuation scores are sparse over the vocabulary, retaining only $K$ candidates per history, but their weighted combination produces a dense vector in the model's embedding space.
\begin{equation}
 \{(w_k,\,S(w_k\mid h_{t,j}))\}_{k=1}^{K}=\mathcal T_K[h_{t,j}],
 \qquad
 f_t=\frac{1}{J}\sum_{j=1}^{J}\sum_{k=1}^{K}
 S(w_k\mid h_{t,j})\,B[w_k].
 \label{eq:fusion}
\end{equation}
We add the COUNT embedding and the token embedding $e_t$ together as the input to the first transformer block, $h_t^{(0)}= f_t + e_t$.

\subsection{N-gram count-based scores}
\label{sec:table-construction}

COUNT can make use of any count-based n-gram LM to score continuations. In our experiments we use \emph{stupid backoff}~\citep{brants2007large} to score observed continuations by their count ratio and otherwise back off to a shorter history with a discount. All of this can be done offline, before pretraining, at a fraction of the cost of training a large LM on the same corpus.

Let $C(h,w)$ count occurrences of a history $h$ followed by token $w$ in the reference corpus, $C(h)$ count the history, and $C(w)$ count the token.
The table stores every history $h$ of the retained lengths with $C(h)\ge c_{\min}$, where $c_{\min}$ is the history-count threshold. \emph{Stupid backoff} scores:
\begin{equation}
 S(w\mid h)=
 \begin{cases}
  C(w)/\sum_{w'\in\mathcal V}C(w'), & h=\varnothing,\\
  C(h,w)/C(h), & h\ne\varnothing,\ C(h,w)>0,\\
  0.4\,S(w\mid h[1:]), & h\ne\varnothing,\ C(h,w)=0.
 \end{cases}
 \label{eq:backoff}
\end{equation}
Here $\varnothing$ is the empty history, whose score is the unigram frequency.
The backoff scores $S(w\mid h)$ need not sum to one and are used without renormalization.

\section{Pretraining experiments}
\label{sec:experiments}

\subsection{Backbone LM}
\label{sec:pretraining}

To represent a standard transformer backbone across a wide range of scales, we use configurations from the \href{https://github.com/allenai/OLMo-core}{OLMo-core} ladder up to 1B parameters and a token to parameter ratio of $D/N=40$, a typical scale for pretraining development experiments. With these configurations we compare standard transformers, COUNT, or learned n-gram embeddings.
We report training budgets as the token-to-parameter ratio $D/N$, where $D$ is the number of training tokens and $N$ counts backbone parameters excluding input embeddings but including the output head.
Our comparisons use $D/N\in\{10,20,40\}$.
We pretrain on \href{https://huggingface.co/datasets/allenai/dolma3_mix-6T-1025-7B}{\nolinkurl{allenai/dolma3_mix-6T-1025-7B}} and use the same corpus to obtain n-gram counts.
Appendix~\ref{app:pretraining} gives the model ladder and optimization settings.

\subsection{Building n-gram tables}
\label{sec:table-building}

We count n-grams through order six over the full Dolma 3~\citep{olmo2025olmo3} pretraining corpus and construct continuation scores using the stupid-backoff procedure in Section~\ref{sec:table-construction}.
To vary table size, we select which n-gram orders to store, prune low-count n-grams, discard histories below a history-count threshold, retain the top-$K$ continuations, and quantize their scores. Since host RAM is inexpensive, our flagship experiments use the largest n-gram table which is sized to saturate the host RAM available on our compute nodes. That 797 GB table serves the 370M, 600M, 760M, and 1B backbones; the 60M, 100M, and 190M backbones use a 17.1 GB table built by the same procedure with tighter count thresholds, so COUNT's table capacity grows with backbone size across the ladder. See Appendix~\ref{app:count-extraction} for all configurations of our primary table.

\subsection{Learned n-gram embeddings}
\label{sec:learned-ngram-embeddings}

There are many approaches to learning n-gram embeddings in the literature (Section~\ref{sec:related-work}); we compare COUNT with Over-Encoding~\citep{huang2025over} as implemented in LongCat~\citep{liu2026scaling} for its released codebase and models, demonstrated at large scales. Over-Encoding augments token embeddings with learned representations of n-gram histories.
Let $N_{\mathrm L}$ be the maximum learned n-gram order.
Each n-gram order $n\in\{2,\ldots,N_{\mathrm L}\}$ uses $R$ hash functions, indexed by $r$.
Each hash selects a row from its own embedding table $T_{n,r}\in\mathbb R^{M_{n,r}\times d_s}$, where $M_{n,r}$ is the number of rows and $d_s$ is the n-gram embedding width.
Learned projections $P_{n,r}\in\mathbb R^{d\times d_s}$ map the retrieved embeddings to model width, giving the n-gram contribution $q_t$ and additively updated transformer input:
\begin{equation}
 q_t=\sum_{n=2}^{N_{\mathrm L}}\sum_{r=1}^{R}
 P_{n,r}\,T_{n,r}\!\left[
   \left(\sum_{j=0}^{n-1}x_{t-j}|\mathcal V|^j\right)\bmod M_{n,r}
 \right],
 \qquad h_t^{(0)}=e_t+q_t.
 \label{eq:over-encoding}
\end{equation}
The tables and projections are learned jointly with the backbone.

LongCat~\citep{liu2026scaling} includes a technique called embedding amplification, which normalizes and scales the combined input, $\sqrt d\,\operatorname{LayerNorm}(h_t^{(0)})$. Here $\operatorname{LayerNorm}$ normalizes across the hidden features with learned gain and bias. We use bigram and trigram embeddings with four hash tables per order and table width $d/8$, projecting each retrieved vector to model width $d$. Following LongCat's guidance, we use a 1:1 ratio of learned input-embedding parameters to backbone parameters: input-token embeddings plus n-gram table parameters equal all remaining learned parameters, including the output head and projections.

We find that embedding amplification also works well with COUNT (Appendix~\ref{app:embedding-amplification}), so we apply it to both COUNT and learned n-gram embeddings.
We also tried training standard transformers with embedding amplification and found it did not improve performance, so we use no embedding amplification for this baseline in our comparisons with COUNT and learned n-gram embeddings.

\subsection{Evaluation}
\label{sec:evaluation}

We summarize OLMo-core's language-modeling and downstream evaluation suites with three aggregates. 
\paragraph{LM validation loss} is the mean token cross-entropy, in nats, across 11 validation sources based on Paloma~\citep{magnusson2024paloma}: C4, Dolma Books, Dolma Common Crawl, Dolma peS2o, Dolma Reddit, Dolma Stack, Dolma Wiki, ICE, M2D2 S2ORC, Pile, and WikiText-103.
\paragraph{Task loss} is the mean bits per byte (BPB) of reference answers from 74 downstream task configurations spanning knowledge and science, code, math and numerical reasoning, commonsense, reading and cloze, and symbolic and pattern reasoning.
Likelihood metrics such as task loss give more stable comparisons of small base LMs than discrete metrics, which are noisier across model sizes and training budgets~\citep{magnusson2025datadecide,heineman2025signal}, so task loss is our primary downstream metric.
\paragraph{Downstream accuracy} is the mean accuracy across the 41 multiple-choice configurations among those tasks, using length-normalized accuracy (candidate log probability divided by character count) where available and ordinary accuracy otherwise.

Appendix~\ref{app:evaluation} specifies all the individual tasks and shows the task-category breakdown. Each source or task configuration receives equal weight.

\section{Pretraining results}
\label{sec:results}

\begin{figure}[]
 \centering
 \includegraphics[width=\linewidth]{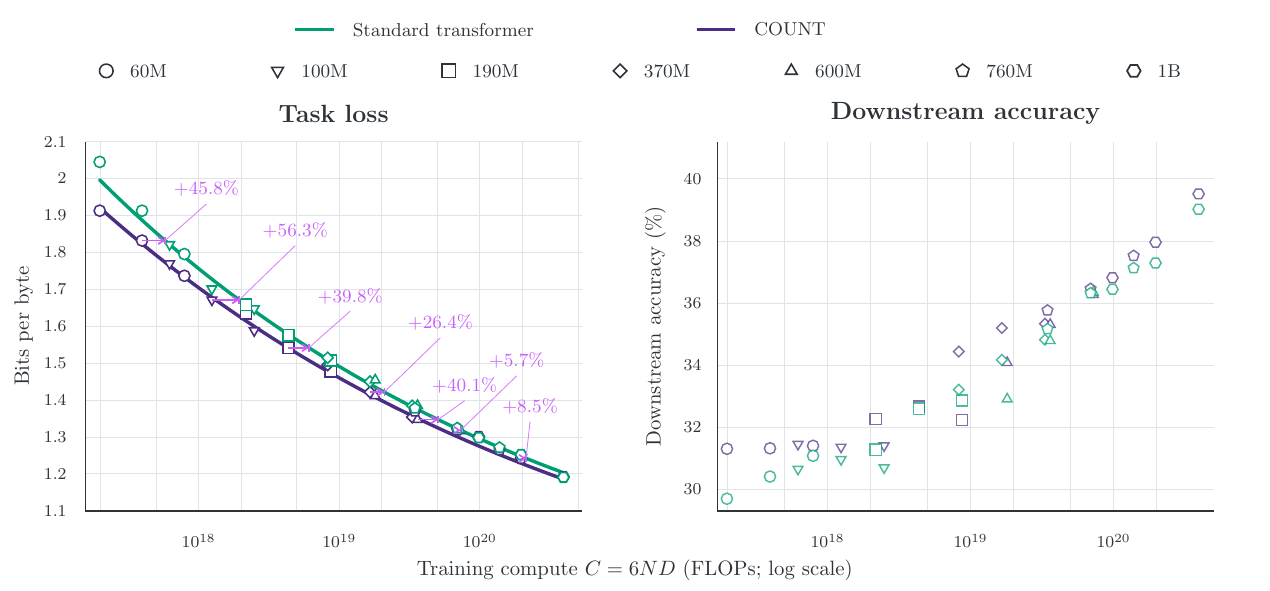}
 \caption{COUNT improves performance without increasing compute or learned parameters. By storing up to 797GB of n-gram frequencies, COUNT allows scaling on host RAM. \textbf{Left:} task loss against training compute at token to parameter ratios $D/N\in\{10,20,40\}$. The solid curves minimize fitted scaling laws for each arm, showing compute optimal performance~\citep{hoffmann2022training}, and violet guides show the additional compute required to make a standard transformer reach the loss of an optimally trained COUNT run. \textbf{Right:} downstream accuracy for the same runs. }
 \label{fig:six-arm-quality}
\end{figure}

\subsection{Scaling host storage saves GPU compute}
\label{sec:unconstrained-memory-results}

Figure~\ref{fig:count-architecture} (right) compares LM loss, and Figure~\ref{fig:six-arm-quality} task loss and downstream accuracy, for standard transformers and COUNT across 60M--1B parameter backbone LMs. COUNT improves downstream accuracy over standard transformer at matched compute at almost all budgets, by storing up to 797GB of n-gram frequencies in host RAM.

We fit loss as a function of backbone size and training tokens separately for each method and loss metric, then select the fitted model/data allocation that minimizes loss at each compute budget~\citep{hoffmann2022training}:
\begin{equation}
 \widehat L(N,D)=E+\frac{A}{N^{\alpha}}+\frac{B}{D^{\beta}},
 \qquad
 \widehat L_{\mathrm{opt}}(C)=\min_{N,D:\,6ND=C}\widehat L(N,D).
 \label{eq:scaling-fit}
\end{equation}
Here $E$, $A$, $B$, $\alpha$, and $\beta$ are all fitted parameters. In our experiments, using COUNT improves LM loss as much as using 46--89\% more compute. Meanwhile, task loss with COUNT improves comparably to a 6--56\% increase in compute.
Like other n-gram memory techniques, COUNT increases performance by scaling memory usage, but it is important to remember that, unlike other methods, COUNT only needs host memory, which is abundant in practice and largely uncontended by other parts of the language modeling pipeline.

\subsection{Using host RAM for n-grams frees GPU RAM for backbone parameters}
\label{sec:finite-memory-results}

\begin{figure}[!t]
 \centering
 \includegraphics[width=\linewidth]{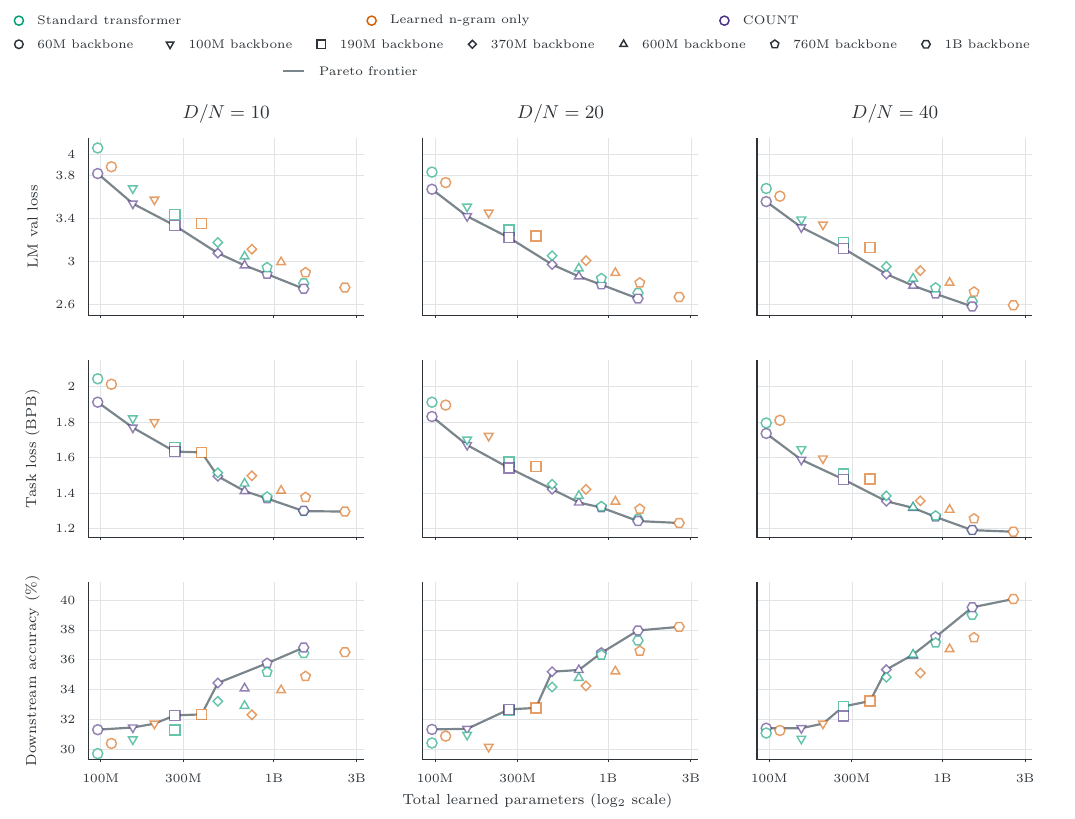}
 \caption{For a given budget of total learned parameters (a hardware-agnostic measure that drives GPU memory footprint) COUNT delivers best LM loss across a range of backbone model scales from 60M--1B, and downstream accuracy (bottom row) follows the same trend with more noise. Each column uses the indicated token to parameter ratio $D/N\in\{10,20,40\}$ relative to that run's own backbone.}
 \label{fig:six-arm-memory}
\end{figure}

Figure~\ref{fig:six-arm-memory} compares standard transformers, COUNT, and learned n-gram embeddings with 60M--1B backbones by total learned parameters. Learned parameters are the primary driver of additional GPU memory footprint, so this comparison allows hardware-agnostic assessment of optimality with respect to memory. The Pareto frontier mostly follows COUNT, because its improvements add no learned parameters. Even where learned n-gram embeddings produce comparable performance, they do so at the cost of additional learned parameters to store learned n-gram embeddings, which cannot be easily offloaded during training and contend with backbone model parameters for GPU memory.
Each column fixes $D/N$, so training tokens scale with the backbone rather than with added embedding parameters.
The Pareto frontier connects configurations with the lowest measured loss, or highest accuracy, for their parameter budget.
The bottom row of Figure~\ref{fig:six-arm-memory} shows downstream accuracy on the same axes; it follows the loss rows with more noise, and the frontier occasionally passes through a standard transformer or learned n-gram embedding point.

Late fusion (Appendix~\ref{app:late-fusion}), which combines count-based next-token distributions with neural predictions at the output, also keeps count tables in host RAM without adding learned n-gram parameters that compete with the backbone for GPU memory. However, constructing input embeddings with COUNT gives larger generalist gains. Also we find that activating late fusion during pretraining does not improve on purely inference-time late fusion. Inference-only late fusion remains a useful option when the pretraining process cannot be changed, but using COUNT at pretraining is best and costs only host RAM.

\section{Inference-time table swapping}
Unlike tables of n-gram embeddings learned from scratch, COUNT tables store scores over a shared vocabulary of next tokens; COUNT maps these scores into embeddings using the same unembedding basis. This allows the n-gram table to be modified at inference time without any training. We test two use cases where table swapping is desirable: whether a pretrained COUNT model can adapt to a new domain-specific table or a smaller table for a tight deployed host-memory budget.
Both experiments use COUNT pretrained on a 1B backbone at token-to-parameter ratios $D/N\in\{10,20,40\}$.

\subsection{Adapting to a new domain without training}
\label{sec:domain-swap-results}

To test an inference-time switch to a domain-specific n-gram count table, we construct math, biomedical, and law tables of 83.398, 81.723, and 80.151 GB, respectively, compared with 797.487 GB for the original general-domain table.
The math table uses 10.66B tokens from the Dolmino Math subset of Olmo 3 midtraining data~\citep{olmo2025olmo3}.
The biomedical and law tables use the health and crime-and-law topic splits of Dolma 3's olmOCR science-PDF source, containing 95.39 and 12.34 billion tokens.
All use the pretraining tokenizer and n-grams through order six; Appendix~\ref{app:domain-tables} gives the construction settings.

For each order, we combine the tables' candidate sets and interpolate their scores as $S_\lambda(w\mid h)=(1-\lambda)S_{\mathrm{general}}(w\mid h)+\lambda S_{\mathrm{domain}}(w\mid h)$, with zero contribution from a table when a candidate is absent from its row. We vary the domain-table weight $\lambda\in\{0,0.25,0.5,1\}$; $\lambda=0$ uses the original table and $\lambda=1$ replaces it completely.
We then select the eight highest-scoring candidates without renormalizing their scores.
We measure task loss and task accuracy on matched domain evaluations: GSM8K, MATH, and Basic Skills arithmetic for math; MedQA, MedMCQA, LabBench, and ten MMLU biomedical subjects for biomedical; and the professional-law, international-law, and jurisprudence subsets of MMLU for law.

\begin{figure}
 \centering
 \includegraphics[width=\linewidth]{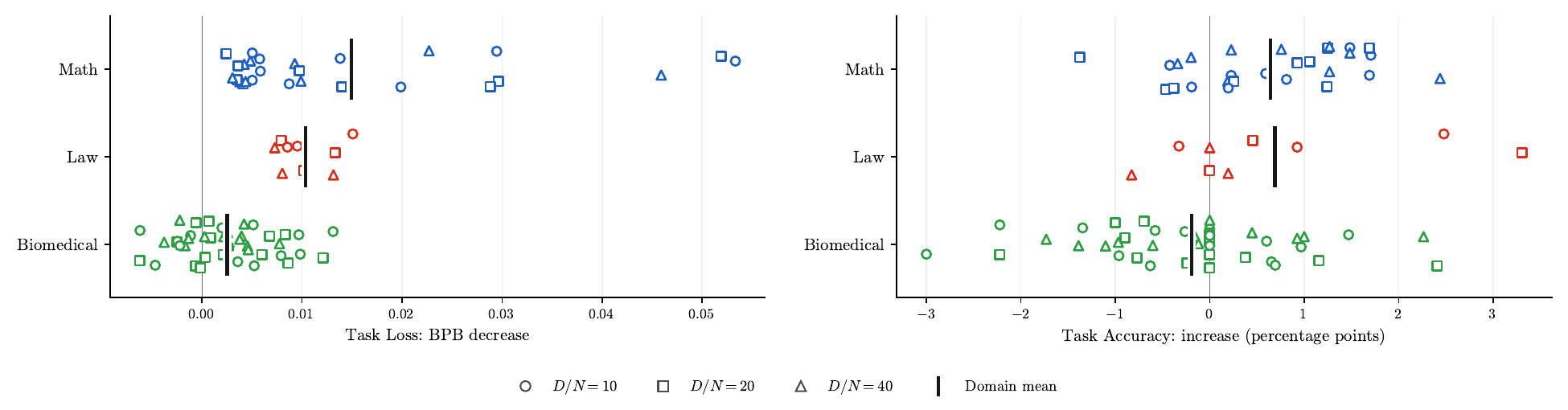}
\caption{The count-based continuation scores in COUNT can be cheaply fit on new domains and swapped in at inference time with no training. We find in-domain task loss improves on all domains when interpolating 1:3 between a domain-specific table and the domain general table used in pretraining this 1B parameter COUNT checkpoint. Accuracy results are more mixed but still increase on average for math and law domains. Each point is a result on one subtask at one training budget. Count-based late-fusion methods also support domain adaptation (Figure~\ref{fig:late-fusion-domains}); COUNT combines this capability with the stronger pretrained improvements of early fusion.}
 \label{fig:domain-table-swap}
\end{figure}

Figure~\ref{fig:domain-table-swap} shows absolute changes in individual task results at $\lambda=0.25$, the interpolation value that makes loss decrease for all 3 domains. These are compared with each task's score with the original table at the same training budget. Mathematical and legal task losses and accuracy improve more than biomedical, which shows mixed results. For reference, at $D/N=20$, COUNT with the generalist table reduces task loss by 0.010 BPB relative to the standard 1B transformer (Figure~\ref{fig:six-arm-quality}), comparable to the improvements from moderate domain weighting. The full interpolation grid is in Appendix~\ref{app:domain-swap-grid}.
Count-based late-fusion methods also support domain adaptation (Figure~\ref{fig:late-fusion-domains} in Appendix~\ref{app:late-fusion}); COUNT combines this capability with the stronger pretrained improvements of early fusion (Figure~\ref{fig:late-fusion-generalist}).

\subsection{Shrinking the table to fit a RAM-constrained deployment}
\label{sec:smaller-table-results}

\begin{figure}[!t]
 \centering
 \includegraphics[width=\linewidth]{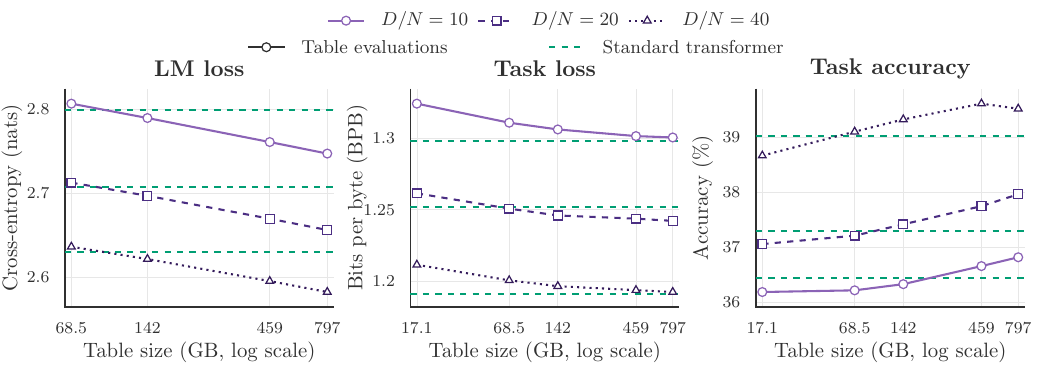}
 \caption{The n-gram table used in COUNT can be replaced at inference time with more compressed versions. This allows the method to work on host-RAM constrained deployments with a graceful and predictable performance degradation.}
 \label{fig:generalist-table-swap-losses}
\end{figure}

The amount of host RAM at training time may be greater than what is available at deployment, so we ask how performance will degrade if we use a smaller n-gram count table at inference time. We replace the same model's original table with general-domain tables of 459.222, 142.037, 68.458, and 17.114 GB at each training budget.
These tables vary in history-count and n-gram-count thresholds.
Figure~\ref{fig:generalist-table-swap-losses} shows how much table storage can be removed while retaining better performance than a standard transformer.
On the 1B backbone COUNT checkpoints we can reduce storage by more than an order of magnitude while still improving on the standard transformer. Notably, this can be done at inference time without any training, as the larger and smaller tables all represent n-gram histories in the same format as next-token scores.

\section{Training and inference throughput}
\label{sec:throughput-results}
\begin{figure}[!t]
 \centering
 \includegraphics[width=\linewidth]{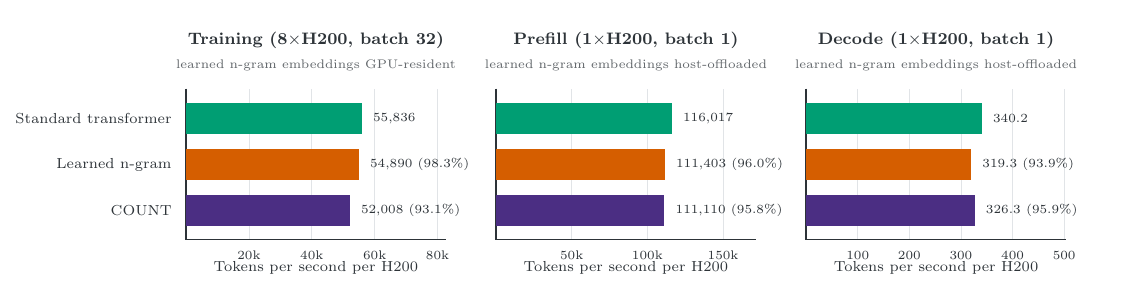}
 \caption{COUNT retains most of the training and inference throughput of a standard transformer. Its throughput is similar to learned n-gram embeddings, and further optimizations would not have far to go to recover both methods to full standard transformer throughput. These measurements use the 1B configurations for each method, and other configurations with smaller tables and lower $J$ further improve COUNT's throughput, see Appendix~\ref{app:throughput-results}.}
 \label{fig:throughput-bars}
\end{figure}

While COUNT adds no learned parameters and can hide the host-resident table lookup time behind backbone LM GPU compute, we still need to measure its costs in practice.
Adding no learned parameters means no added cost in the $C=6ND$ sense and in parameter memory, but gathering selected unembedding rows is real work that the estimate omits and GPU footprint depends on more than parameters, so we measure throughput (Figure~\ref{fig:throughput-bars}) and peak GPU memory (Table~\ref{tab:throughput-memory}) directly.
These experiments use the 1B backbone parameter checkpoints of COUNT, learned n-gram embeddings, and standard transformer.
We measure training throughput as median tokens per second (TPS) of optimization steps 50-100 on 8 H200 GPUs.
For inference we measure prefill and decode on one H200 after warm-up batches. Appendix~\ref{app:throughput} gives the measurement configurations and Appendix~\ref{app:throughput-results} reports additional measured settings.

COUNT has a small impact on throughput at the 1B backbone scale, retaining 93.1--97.8\% of standard transformer training throughput, 95.5--95.8\% for prompt prefill, and 95.9--96.7\% for autoregressive decoding.
We use three optimizations to recover most of the standard transformer's throughput: (1) compiled, multithreaded host lookup; (2) prefetching lookups for training and prefill tokens while the GPU processes the preceding batch; and (3) a fused embedding reduction that gathers and sums candidate vectors without storing them separately.

We also optimize the learned n-gram embedding baseline that we compare against: (1) we implement sparse optimizer updates for only the retrieved embedding rows, avoiding a full-table gradient update at each step.
At 190M with a 17.1 GB learned n-gram table, full-table optimizer updates ran at 4.4\% of standard transformer training throughput and updates to retrieved rows only at 96.3\%, and the 1B learned baseline in Table~\ref{tab:throughput-training} trains at 98.3\%.
(2) We implement inference-time host offloading to gather learned embedding rows on the CPU and transfer them to the GPU, retaining 95.9--96.0\% of standard transformer prefill throughput and 93.9--96.0\% of decode throughput, similar to COUNT. 

\section{Related work}
\label{sec:related-work}

\subsection{N-grams in neural language models}
There has been much work on learning embeddings for n-grams, especially in the past two years, with the motivation that scaling embeddings increases performance predictably \citep{tao2024scaling}.
The core problem all such work grapples with is that independently representing all $|\mathcal V|^n$ possible n-grams is quickly intractable.
One line of work uses multiple hashes of the n-grams to allow collisions that partially cancel out over different hash functions~\citep{roy2022ngrammer,huang2025over,liu2026scaling}.
Other work builds embeddings of an n-gram based on learned representations of the tokens that it shares with lower n-gram orders~\citep{zhou2026tensorizing} or combines token embeddings that make up the current n-gram suffix at inference time without any training~\citep{qu2026ngm}.
Yet another approach is to use an auxiliary model to embed the n-grams~\citep{yu2025scaling}.
Similarly, the backbone model's own hidden state can provide the contextual information, and additional embeddings are keyed just by the current unigram but further conditioned by the hidden state~\citep{ding2026meki,tseng2026large}.
This approach can also be used with hashed n-gram embeddings by having the hidden state of the model gate the inclusion of the embeddings~\citep{cheng2026engram}.

Most similar to our work, \citet{neubig2016generalizing} actually mix count-based next-token distributions with neural predictions conditioned on some count-based n-gram features.
There is also more recent late-fusion-only work that interpolates the neural LM's output distribution with a nonparametric one at inference time, from nearest neighbors over a datastore of contexts~\citep{khandelwal2020knnlm} or from unbounded-order Infini-gram counts~\citep{liu2024infinigram}.
\citet{li2022residual} instead train a neural LM with scaled n-gram log probabilities added to its output logits, yielding a multiplicative combination of the two distributions.
This work on late fusion of next-token frequencies also demonstrates domain adaptation at inference time~\citep{khandelwal2020knnlm,li2022residual}.
In contrast to these approaches, our method can be seen as an early fusion combination of count-based and neural LMs where the count-based next-token distribution is represented directly in the shared embedding space of the neural LM.
We compare late fusion to COUNT in Appendix~\ref{app:late-fusion} and find that late fusion does not benefit from inclusion in pretraining and does not improve generalist performance as much as COUNT.

Outside of academic literature, nanoGPT speed-run submissions concurrent with COUNT (\href{https://github.com/KellerJordan/modded-nanogpt/pull/367}{PR \#367} and \href{https://github.com/KellerJordan/modded-nanogpt/pull/380}{PR \#380}) and unpublished experiments discussed in \href{https://liujch1998.github.io/2025/08/02/llm-data.html}{a blog post} by Infini-gram author Jiacheng Liu retrieve a sample of matching suffixes from training data and use their continuations to build a next-token hint from the token embeddings, which is then input to the LM backbone. Where these methods predict continuations from less than a hundred samples retrieved online, COUNT computes corpus frequencies over all occurrences in a full pretraining corpus offline. We compare also the use of the input token embedding or unembedding matrix as next-token basis in Appendix~\ref{app:count-basis} and find that our use of the unembedding performs best.
These engineering experiments highlight further opportunities for throughput optimization of count-based embeddings, complementing our systematic analysis across 60M--1B backbone scales, training budgets, and 74 downstream task evaluations.

\subsection{Tokenization and n-grams}
A highly related thread of work explores n-gram tokens for a different motivation.
They address the arbitrariness of token segmentation by either exploring ways to segment further and then represent combinations of tokens~\citep{clark2022canine,pagnoni2025byte} or by relaxing segmentation and allowing n-grams composed of many words to become single LM tokens~\citep{liu2025superbpe}.

\subsection{Retrieval-augmented language models}
Though architecturally unrelated, retrieval-augmented LMs~\citep{guu2020retrieval,lewis2020retrieval,borgeaud2022improving,asai2024selfrag,shao2024scaling} share a motivation with COUNT: access to nonparametric information from a raw corpus can complement the parametric knowledge learned by the LM when it trains on the same data.
Notably, retrieval also provides the affordance of zero-training domain adaptation.
Where these methods practically differ from COUNT is that their nonparametric information is input directly as a token sequence into the LM, increasing context usage and compute.
Also because COUNT stores only counts, its data store will always be much smaller than storing the raw corpus it is fit on.

\section{Conclusion}
High-bandwidth GPU memory is the scarcest and most expensive resource in current training and serving hardware, while host RAM is cheaper per byte and largely uncontended by the model itself.
Learned n-gram embeddings buy their gains with parameters and optimizer state that must sit in GPU memory alongside the backbone.
COUNT buys comparable gains with bytes on cheaper memory, leaving GPU memory for backbone parameters.
Representing n-grams by counts also opens possibilities beyond efficiency.
A table of counts is an aggregate statistic rather than a corpus, so sharing it is often less sensitive and more compact than sharing the documents it summarizes, which offers a new way to ingest data that cannot be released as text.
N-gram frequencies can also be estimated from far less data than pretraining requires.
With only a small amount of domain-specific data we fit count-based priors that steered a generalist, pretrained model to better in-domain performance.
The long history of settled knowledge on count-based language modeling can likely produce useful next-token priors on even smaller corpora that are too small for parametric learning.
Finally, because each row lists the next-token candidates and scores for a history, the prior a COUNT model consults can be read, corrected, or extended directly, by editing rows rather than by retraining.

\subsection*{AI use statement}

In this work, we used generative AI tools for implementing methods. We have not used generative AI tools for other tasks with required disclosure (design or provide feedback on research methodology or experiments, clean and reformat dataset, support qualitative and thematic data analysis, interpret results, propose or refine hypotheses), and the rest of the disclosure tasks (generate synthetic data sets, help develop theoretical models or conceptual frameworks, formulate mathematical claims, provide critical ingredients for proving mathematical claims, assist in the writing of proofs, assist with translation) are not applicable to this work. Additionally, we used generative AI tools to create or modify scientific figures or images, suggest experimental parameters, create or edit software code, draft parts of a research paper, edit a research paper to improve readability, identify relevant literature, format references, and propose keywords for a research paper. We have reviewed all AI-assisted work. The paper was fully revised multiple times by all authors. The final method implementation was manually reviewed by the first author. We take responsibility for the final content of this work, including text, claims or artifacts produced with the aid of generative AI.

\subsection*{Ethics statement}

This work aims to make language models more efficient in order to lower the cost and environmental impact of building and running them. By moving n-gram memory from GPU memory to host RAM and adding no learned parameters, COUNT lets smaller models reach quality that would otherwise require more compute or more expensive hardware, and its tables can be built from modest corpora. We hope this helps a wider range of people and organizations, including those with limited compute or limited data, to train and use capable models. COUNT also changes what a data contributor has to share: an n-gram count table records how often continuations follow short contexts, not the documents themselves, and is therefore less sensitive than raw pretraining text, although high-order counts over rare contexts can still reveal fragments of the source and should be released with the same care as other derived statistics. Our experiments use publicly available pretraining data and evaluation suites.

\subsection*{Reproducibility statement}

Section~\ref{sec:method} defines COUNT and the stupid-backoff scoring it uses, and Section~\ref{sec:learned-ngram-embeddings} defines the learned n-gram baseline in the same notation.
Every pretraining run follows the OLMo-core ladder and recipe in Appendix~\ref{app:pretraining}; the primary n-gram table's construction is stated in full in Appendix~\ref{app:count-extraction}, together with the smaller general-domain tables and the domain-specific tables in Table~\ref{tab:domain-tables}.
Section~\ref{sec:evaluation} and Appendix~\ref{app:evaluation} specify the evaluation suite, metrics, and aggregations, and Appendix~\ref{app:throughput} records the hardware, batch sizes, and timing procedure behind every throughput number, with the complete measurements in Appendix~\ref{app:throughput-results}.
Appendix~\ref{app:results} reports the additional results behind the ablations and inference-time sweeps.
We release the n-gram counts extracted from the full Dolma 3 pretraining corpus, the table-building and training code, and pretrained checkpoints for COUNT and the learned n-gram baselines.

\subsection*{Acknowledgments}
We would like to thank Will Merrill, Jiacheng Liu, and Shannon Shen for fruitful discussions on combining n-gram and neural LMs, and Yashas Samaga for discussions on model ladder experiments. IM is supported by the NSF CSGrad4US Fellowship. This material is based upon work supported by the U.S. National Science Foundation under Award Nos. 2313998 and 2413244. Any opinions, findings, and conclusions or recommendations expressed in this material are those of the author(s) and do not necessarily reflect the views of the U.S. National Science Foundation. This work was supported by the Singapore National Research Foundation and the National AI Group in the Singapore Ministry of Digital Development and Information under the AI Visiting Professorship Programme (award number AIVP-2024-001) and the AI2050 program at Schmidt Sciences.

\bibliography{references}
\bibliographystyle{iclr2027_conference}

\clearpage
\appendix
\section{Experimental details}
\label{app:configs}

\subsection{OLMo-core}
\label{app:pretraining}

Table~\ref{tab:model-ladder} lists the backbone geometries and Table~\ref{tab:training-config} the training recipe. All arms use the same backbone at each ladder size; size labels are shorthand rather than exact parameter counts. For example, the 1B configuration has $N=1{,}279{,}395{,}840$ parameters under our budget convention. Token budgets use this exact parameter count, excluding input embeddings and added n-gram parameters but including the output head, and are rounded to full optimizer steps.

\begin{table}[htbp]
 \centering
\caption{OLMo-core model ladder configurations.
 The feed-forward width is the intermediate width of each transformer block.}
 \label{tab:model-ladder}
 \begin{tabular}{lrrrr}
  \toprule
  Ladder label & Layers & Hidden width & Heads & Feed-forward width \\
  \midrule
  60M  &  8 &   384 &  8 & 1,536 \\
  100M & 12 &   512 &  8 & 2,048 \\
  190M & 12 &   768 & 12 & 3,072 \\
  370M & 16 & 1,024 & 16 & 4,096 \\
  600M & 16 & 1,280 & 16 & 5,120 \\
  760M & 16 & 1,536 & 16 & 6,144 \\
  1B   & 16 & 2,048 & 16 & 8,192 \\
  \bottomrule
 \end{tabular}
\end{table}

\begin{table}[htbp]
 \centering
 \caption{Training configuration.
 $N$ is the backbone parameter count defined above.
}
 \label{tab:training-config}
 \renewcommand{\arraystretch}{1.18}
 \begin{tabularx}{\linewidth}{@{}p{.25\linewidth}X@{}}
  \toprule
  Setting & Value \\
  \midrule
  Tokenizer & Dolma 3 tokenizer (\href{https://huggingface.co/allenai/dolma2-tokenizer}{\nolinkurl{allenai/dolma2-tokenizer}}) \\
  Vocabulary & 100,278 tokens; embedding allocation padded to 100,352 rows \\
  Sequence length & 8,192 tokens \\
  Position encoding & Rotary position embeddings with base 500,000 \\
  Precision & Bfloat16 model computation; float32 gradient reduction \\
  Optimizer & AdamW with anomalous-update skipping (\texttt{SkipStepAdamW}); $\beta_1=0.9$; $\beta_2=0.99$ for 60M and 100M and $0.95$ for 190M--1B; $\epsilon=10^{-8}$ \\
  Weight decay & 0.1; input embeddings, embedding-amplification parameters, and learned n-gram tables and projections are exempt \\
  Peak learning rate & $0.00235(N/108{,}000{,}000)^{-1/3}$ \\
  Schedule & Warmup over $N$ tokens, followed by a stable learning rate and linear decay to zero over the final 10\% of each successive reporting interval (WSD-S) \\
  Token budgets & Reporting endpoints at $D/N=10$, $20$, and $40$ \\
  Global batch & 262,144 (60M and 100M), 524,288 (190M), 786,432 (370M), 983,040 (600M), 1,310,720 (760M), and 1,572,864 (1B) \\
  COUNT & Unembedding matrix $B$ as the next-token basis; tables store $K=32$ candidates per history and training uses $K=8$ and $J=6$ with embedding amplification; 17.1 GB table at 60M--190M and 797.5 GB at 370M--1B \\
  Learned n-gram embeddings & Orders two and three, four hash tables per order, embedding width $d/8$ projected to $d$; rows set by the 1:1 ratio: 0.40M rows across the eight tables at 60M, 0.79M at 100M, 1.19M at 190M, 2.11M at 370M, 2.63M at 600M, 3.16M at 760M, and 4.21M at 1B (0.04--2.16 GB of bfloat16 table weights). The table weights are updated with sparse row-wise AdamW on retrieved rows, using the backbone learning-rate schedule and betas, with zero weight decay \\
  \bottomrule
 \end{tabularx}
\end{table}

\subsection{N-gram tables}
\label{app:count-extraction}
\label{app:domain-tables}
\label{app:math-tables}
We build smaller tables from intermediate artifacts, which have their own count thresholds and limits on the number of continuation candidates retained per history.

Our primary general-domain table stores the top $K=32$ continuations for every history of order one through six seen at least 265 times, after dropping order-4 to order-6 n-grams with fewer than seven occurrences, and occupies 797.487 GB.

\begin{figure}
 \centering
 \includegraphics[width=\linewidth]{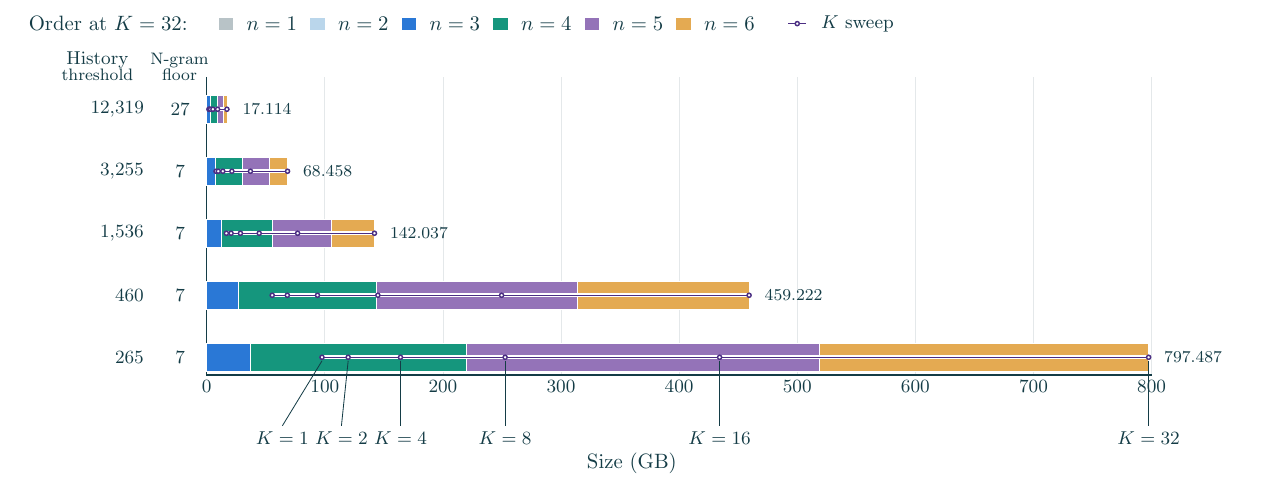}
 \caption{The size of our n-gram tables depends on several configurations. History and n-gram occurrence thresholds limit the number of rows in the table. Each order of n-gram is its own set of rows in the table, with orders four through six accounting for most of the rows. Finally lowering K to store fewer top continuations reduces the table width.}
 \label{fig:table-size}
\end{figure}

Figure~\ref{fig:table-size} shows how the general-domain table sizes depend on the n-gram order, the history-count threshold, the per-n-gram count floor, and $K$; sizes are decimal gigabytes including headers.
For storage, we quantize $\log_{10} S(w\mid h)$ to 256 levels per order, spanning the smallest retained finite log score through zero, and convert the decoded values to natural-log scores at read time.

The domain tables in Table~\ref{tab:domain-tables} are built from the Dolmino Math subset of Dolma 3 (10.66B tokens) and from the Health and Crime and Law topics of Dolma 3's olmOCR science-PDF source (95.39B and 12.34B tokens), following the WebOrganizer topic categories~\citep{wettig2025organize,olmo2025olmo3}.
All use the pretraining tokenizer, orders through six, a count floor of one, and a backoff factor of $0.4$; the biomedical and law history-count thresholds are the lowest integers that fit within the math table's 83.398 GB.

\begin{table}[htbp]
 \centering
 \caption{Domain-specific tables used in Figure~\ref{fig:domain-table-swap}. Sizes are decimal GB with 8-bit scores.}
 \label{tab:domain-tables}
 \label{tab:math-tables}
 \begin{tabular}{lrrr}
  \toprule
  Domain & Size (GB) & $K$ & \shortstack{History-count\\threshold} \\
  \midrule
  Math &  83.398 &  32 & 4 \\
  Biomedical & 81.723 & 32 & 45 \\
  Law & 80.151 & 32 & 8 \\
  \bottomrule
 \end{tabular}
\end{table}

\subsection{Evaluation breakdown}
\label{app:evaluation}

\begin{figure}[htbp]
 \centering
 \includegraphics[width=\linewidth]{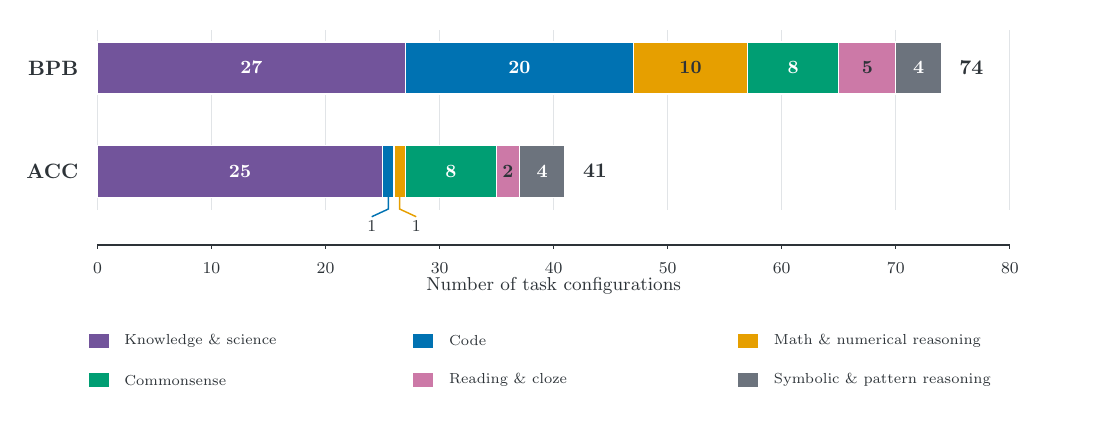}
 \caption{Composition of the task-loss (BPB) and accuracy (ACC) aggregates by qualitative task category. Each eligible configuration contributes once: 74 configurations for BPB and 41 for accuracy. Subject subsets, splits, prompt formats, and language variants are counted separately. Categories describe task families rather than isolated skills.}
 \label{fig:task-aggregate-composition}
\end{figure}

Our downstream evaluations use OLMES task configurations~\citep{gu-etal-2025-olmes} through \texttt{ai2-olmo-eval} 0.9.0, except for the GSM8K and MATH accuracy evaluations described below.
LM validation loss averages token cross-entropy within each of the 11 sources before averaging sources.
Task loss averages the 74 per-configuration BPB values, which include reference-answer scoring for code, math and open-ended tasks that report no accuracy; the accuracy aggregate covers the 41 multiple-choice configurations.
Figure~\ref{fig:task-aggregate-composition} groups the configurations into task categories; subject subsets, splits, prompt formats and language variants count separately, and every configuration keeps equal weight.
The configurations derive from the following parent benchmarks.
\begin{itemize}
 \item \textbf{Knowledge and science:} ARC-Easy and ARC-Challenge~\citep{clark2018arc}; MMLU~\citep{hendrycks2021mmlu}, as its four subject groups on the validation and test splits; LAB-Bench DbQA and ProtocolQA~\citep{laurent2024labbench}; MedQA~\citep{jin2021medqa}; MedMCQA~\citep{pal2022medmcqa}; Natural Questions~\citep{kwiatkowski2019naturalquestions}; Jeopardy~\citep{gu-etal-2025-olmes}; Basic Skills common knowledge~\citep{olmo2025olmo3}.
 \item \textbf{Code:} HumanEval~\citep{chen2021humaneval}; MBPP~\citep{austin2021mbpp}; MT-MBPP, machine-translated MBPP in 17 further programming languages~\citep{olmo2025olmo3}; Basic Skills coding~\citep{olmo2025olmo3}.
 \item \textbf{Math and numerical reasoning:} GSM8K~\citep{cobbe2021gsm8k}; MATH~\citep{hendrycks2021math} in the Minerva format~\citep{lewkowycz2022minerva}, as seven subject subsets; DROP~\citep{dua2019drop}; Basic Skills arithmetic~\citep{olmo2025olmo3}. GSM8K and MATH accuracy in Figure~\ref{fig:domain-table-swap} uses the four-choice versions of these test sets, whose distractor answers were produced by \citet{zhang2024mc}; the model ranks the answer texts as for every other accuracy task.
 \item \textbf{Commonsense:} CommonsenseQA~\citep{talmor2019commonsenseqa}; HellaSwag~\citep{zellers2019hellaswag}; PIQA~\citep{bisk2020piqa}; SocialIQA~\citep{sap2019socialiqa}; WinoGrande~\citep{sakaguchi2021winogrande}.
 \item \textbf{Reading and cloze:} CoQA~\citep{reddy2019coqa}; LAMBADA~\citep{paperno2016lambada}; SQuAD~\citep{rajpurkar2016squad}; QASPER yes/no~\citep{dasigi2021qasper}; SciRIFF yes/no~\citep{wadden-etal-2025-sciriff}.
 \item \textbf{Symbolic and pattern reasoning:} Basic Skills logical reasoning, pattern recognition and string operations~\citep{olmo2025olmo3}; CopyColors~\citep{gu-etal-2025-olmes}.
\end{itemize}
ARC, MMLU, CommonsenseQA, PIQA and SocialIQA each use both prompt formats, ranked-choice (\texttt{mc}) and completion (\texttt{rc}); every other benchmark uses its completion format only.

\subsection{Throughput configurations}
\label{app:throughput}

All throughput measurements use the 1B ladder configuration, include lookup and host--GPU transfer in the timed workload, use the 1:1 input-embedding-to-backbone parameter ratio for the learned n-gram baseline, keep COUNT tables in host RAM, and (for inference) use $D/N=40$ checkpoints on C4 validation prompts.
Compiled lookup searches sorted histories with multiple CPU threads; lookup caching reuses retrieved results; prefetch prepares lookups while the GPU processes the preceding batch; fused embedding combines row gathering and weighted summation; sparse updates optimize only retrieved learned n-gram rows; host offloading keeps learned n-gram tables on the CPU with their projections on the GPU; KV caching and CUDA graphs are the usual inference optimizations.
Tables~\ref{tab:training-throughput-config}--\ref{tab:decode-throughput-config} give the settings for each workload.

\begin{table}[H]
 \centering
 \caption{Training TPS experiment configuration.}
 \label{tab:training-throughput-config}
 \begin{tabularx}{\linewidth}{@{}p{.26\linewidth}X@{}}
  \toprule
  Setting & Value \\
  \midrule
  Hardware & Eight H200 GPUs \\
  Batch per GPU & Four sequences; six gradient-accumulation steps \\
  COUNT table & 17.1 or 797.5 GB \\
  COUNT selection & $K=8$; $J=6$ or $1$ \\
  Learned table & GPU-resident \\
  Optimizations & Prefetch; fused embedding; sparse updates \\
  Runtime & FlashAttention 3; block/head compilation; fully sharded data parallelism \\
  Timing & 100 steps; median of six TPS samples at steps 50--100 \\
  \bottomrule
 \end{tabularx}
\end{table}

\begin{table}[H]
 \centering
 \caption{Prefill TPS experiment configuration.}
 \label{tab:prefill-throughput-config}
 \begin{tabularx}{\linewidth}{@{}p{.26\linewidth}X@{}}
  \toprule
  Setting & Value \\
  \midrule
  Hardware & One H200 GPU \\
  Batch sizes & 1 and 8 sequences \\
  Prompt length & 8,192 tokens \\
  COUNT table & 797.5 GB; $K=8$ per order, orders 1--6 \\
  Learned table & 2.008 GiB; GPU-resident or host offloading \\
  Optimizations & Compiled lookup; lookup caching; prefetch; fused embedding; KV caching \\
  Lookup workers & Eight workers; ten batches of lookahead \\
  Runtime & PyTorch attention \\
  Timing & Three warmup batches; 96 timed batches at batch 1 or 12 at batch 8 \\
  \bottomrule
 \end{tabularx}
\end{table}

\begin{table}[H]
 \centering
 \caption{Decode TPS experiment configuration. Prompt processing and graph capture are excluded from the timed generation steps.}
 \label{tab:decode-throughput-config}
 \begin{tabularx}{\linewidth}{@{}p{.26\linewidth}X@{}}
  \toprule
  Setting & Value \\
  \midrule
  Hardware & One H200 GPU \\
  Batch sizes & 1 and 256 sequences \\
  Prompt / generation & 512 prompt tokens; 256 greedily generated tokens \\
  COUNT table & 797.5 GB; $K=8$ per order, orders 1--6 \\
  Learned table & 2.008 GiB; GPU-resident or host offloading \\
  Optimizations & Compiled lookup; lookup caching; fused embedding; KV caching; CUDA graphs \\
  Lookup threads & Eight \\
  Runtime & PyTorch attention \\
  Timing & Ten warmup steps; 246 timed generation steps \\
  \bottomrule
 \end{tabularx}
\end{table}

\section{Additional results}
\label{app:results}

\subsection{Late fusion}
\label{app:late-fusion}

We compare COUNT with two output-side fusion variants using the same tables and $K=8$, but only the longest matching history $h_{t,1}$ rather than averaging orders, as the neural LM cannot correct noisy low order predictions in late fusion.
Let $p_t(w)=p_\theta(w\mid x_{1:t})$ be the standard transformer's next-token distribution and $\mathcal W_t$ the candidates returned by $\mathcal T_K[h_{t,1}]$, with scores $S(w\mid h_{t,1})$ from Equation~\ref{eq:backoff}.
Write $A_t=\sum_{w\in\mathcal W_t}S(w\mid h_{t,1})$ for their retained mass.
\emph{Additive fusion} follows the probability interpolation used by kNN-LM~\citep{khandelwal2020knnlm} and infini-gram~\citep{liu2024infinigram}, adapted here to top-$K$ stupid-backoff scores by capping the retained mass at one:
\begin{equation}
 \begin{gathered}
 q_t(w)=\frac{S(w\mid h_{t,1})}{\max\{1,A_t\}},\qquad
 m_t=\min\{A_t,1\},\\
 p_t^{\mathrm{add}}(w)=(1-\lambda m_t)p_t(w)+\lambda q_t(w).
 \end{gathered}
 \label{eq:late-additive}
\end{equation}
\emph{Multiplicative fusion} instead keeps at least $\varepsilon=10^{-4}$ total mass for non-candidates, distributed uniformly:
\begin{equation}
 \begin{gathered}
 \widetilde m_t=\min\{A_t,1-\varepsilon\},\\
 \widehat q_t(w)=
 \begin{cases}
  \dfrac{S(w\mid h_{t,1})}{\max\{1,A_t/(1-\varepsilon)\}}, & w\in\mathcal W_t,\\[6pt]
  \dfrac{1-\widetilde m_t}{|\mathcal V|-|\mathcal W_t|}, & w\notin\mathcal W_t.
 \end{cases}
 \end{gathered}
 \label{eq:late-ngram}
\end{equation}
Following the logit-level combination of \citet{li2022residual}, the fused distribution is the normalized product
\begin{equation}
 p_t^{\mathrm{mult}}(w)=
 \frac{p_t(w)\,\widehat q_t(w)^\alpha}
 {\sum_{v\in\mathcal V}p_t(v)\,\widehat q_t(v)^\alpha}.
 \label{eq:late-multiplicative}
\end{equation}
For the inference-only comparisons below, we freeze the standard transformer without embedding amplification and fix $\lambda=0.1$ or $\alpha=0.1$.
Domain adaptation uses the domain table directly; its changes are measured against the same unfused checkpoint, while COUNT's changes are relative to its own generalist-table checkpoint.

\begin{figure}[htbp]
 \centering
 \includegraphics[width=\linewidth]{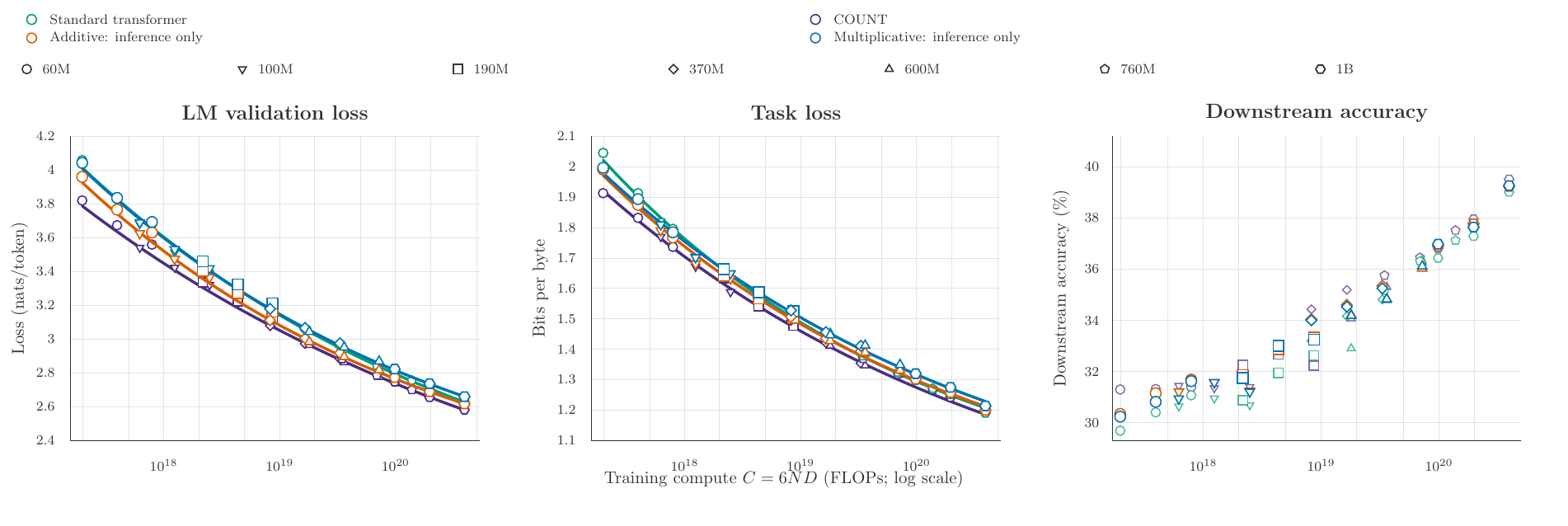}
 \caption{Generalist inference-only late fusion compared with COUNT and the standard transformer.}
 \label{fig:late-fusion-generalist}
\end{figure}

\begin{figure}[htbp]
 \centering
 \includegraphics[width=\linewidth]{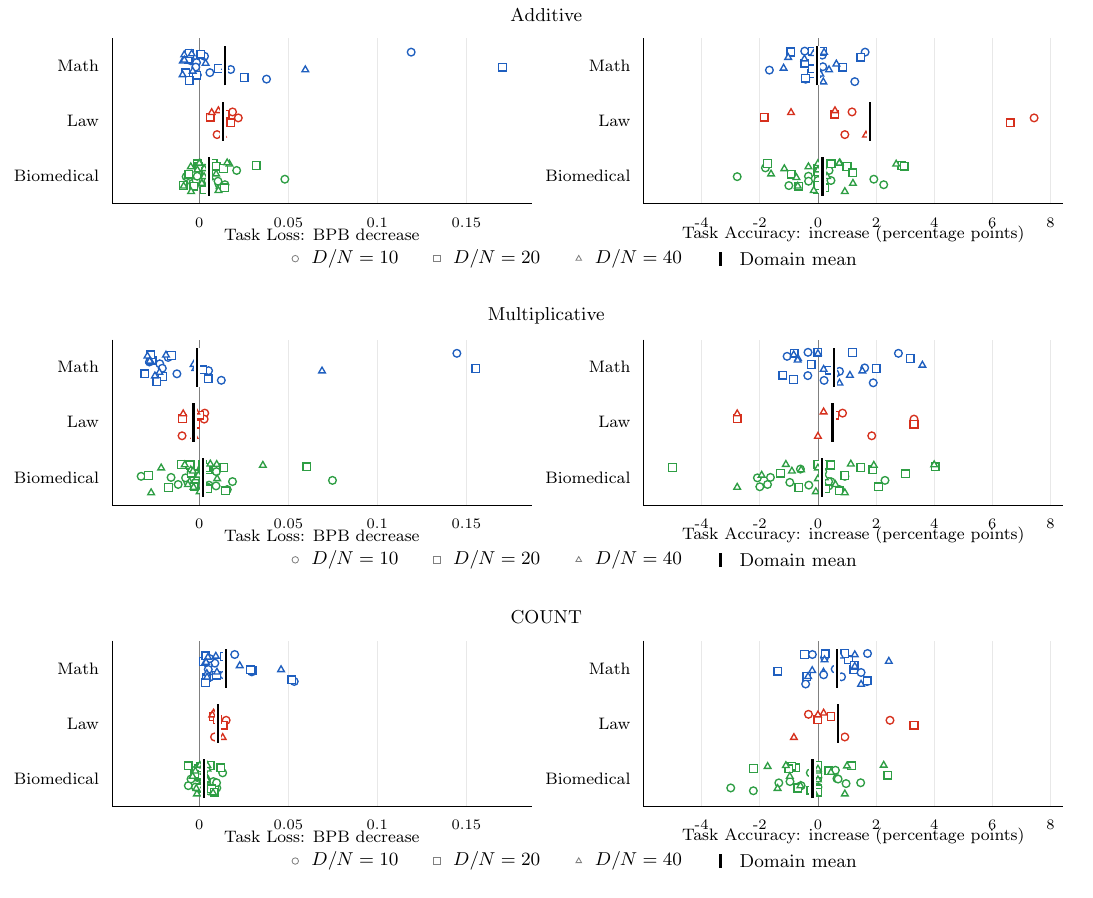}
 \caption{Domain adaptation with inference-only late fusion and COUNT.}
 \label{fig:late-fusion-domains}
\end{figure}

Figure~\ref{fig:late-fusion-generalist} shows that COUNT achieves better generalist performance than either additive or multiplicative late fusion.
Figure~\ref{fig:late-fusion-domains} shows that training-free adaptation is an affordance shared by count-based methods; COUNT combines this capability with better pretrained generalist performance.

\begin{figure}[htbp]
 \centering
 \includegraphics[width=\linewidth]{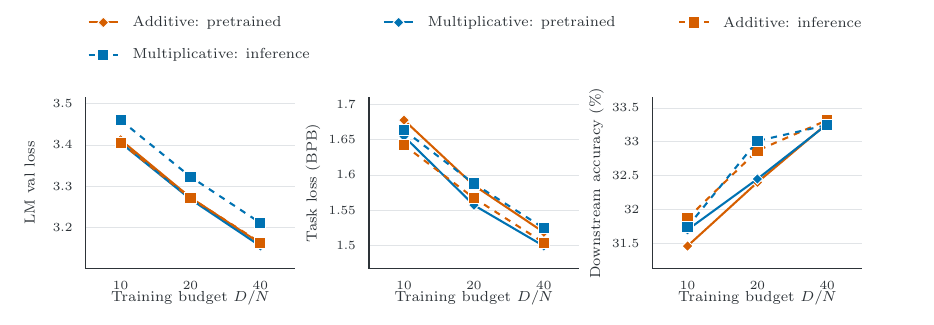}
 \caption{Pretraining versus inference-time late fusion at 190M.}
 \label{fig:late-fusion-pretraining}
\end{figure}

\paragraph{Pretraining versus inference-time fusion.}
At 190M, we also pretrain each late-fusion arm with its scalar weight learned from an initialization of $0.1$, using the same 17GB generalist table as its inference-only counterpart.
Figure~\ref{fig:late-fusion-pretraining} shows that pretraining with late fusion does not improve performance over applying late fusion only at inference time, so we use inference-only fusion at the other scales.

\subsection{Full throughput results}
\label{app:throughput-results}

Tables~\ref{tab:throughput-training}--\ref{tab:throughput-decode} report every measured setting behind Figure~\ref{fig:throughput-bars}.
Rates are tokens per second per executing H200; parentheses give each treatment's share of the standard transformer.
Table~\ref{tab:throughput-memory} reports the peak GPU memory of the same measurements.
The learned baseline's training rate is a single measurement.
Placement refers to the additional n-gram storage: COUNT's count tables are host-resident in all measurements, while the backbone's unembedding matrix is GPU-resident for every method.

\begin{table}[htbp]
 \centering
 \setlength{\tabcolsep}{4pt}
 \caption{Training throughput at batch 32 on eight H200s by COUNT lookup ($J$) and table size.}
 \label{tab:throughput-training}
  \label{tab:throughput}
 \begin{tabular}{@{}lrrrr@{}}
\toprule
 & \shortstack{$J=6$\\17.1 GB} & \shortstack{$J=1$\\17.1 GB} & \shortstack{$J=6$\\797.5 GB} & \shortstack{$J=1$\\797.5 GB} \\
\midrule
Standard transformer & 55,836 & 55,836 & 55,836 & 55,836 \\
Learned n-gram & 54,890 (98.3\%) & 54,890 (98.3\%) & 54,890 (98.3\%) & 54,890 (98.3\%) \\
COUNT & 52,127 (93.4\%) & 54,585 (97.8\%) & 52,008 (93.1\%) & 54,532 (97.7\%) \\
\bottomrule
\end{tabular}

\end{table}

\begin{table}[htbp]
 \centering
 \setlength{\tabcolsep}{4pt}
 \caption{Prefill throughput on one H200 by batch size and n-gram storage placement.}
 \label{tab:throughput-prefill}
 \begin{tabular}{@{}lrrrr@{}}
\toprule
 & \shortstack{Batch 1\\GPU-resident} & \shortstack{Batch 8\\GPU-resident} & \shortstack{Batch 1\\Host-resident} & \shortstack{Batch 8\\Host-resident} \\
\midrule
Standard transformer & 116,017 & 122,389 & 116,017 & 122,389 \\
Learned n-gram & 115,181 (99.3\%) & 122,091 (99.8\%) & 111,403 (96.0\%) & 117,402 (95.9\%) \\
COUNT & --- & --- & 111,110 (95.8\%) & 116,931 (95.5\%) \\
\bottomrule
\end{tabular}

\end{table}

\begin{table}[htbp]
 \centering
 \setlength{\tabcolsep}{4pt}
 \caption{Decode throughput on one H200 with CUDA graphs by batch size and n-gram storage placement.}
 \label{tab:throughput-decode}
 \begin{tabular}{@{}lrrrr@{}}
\toprule
 & \shortstack{Batch 1\\GPU-resident} & \shortstack{Batch 256\\GPU-resident} & \shortstack{Batch 1\\Host-resident} & \shortstack{Batch 256\\Host-resident} \\
\midrule
Standard transformer & 340.2 & 19,216 & 340.2 & 19,216 \\
Learned n-gram & 290.8 (85.5\%) & 18,795 (97.8\%) & 319.3 (93.9\%) & 18,448 (96.0\%) \\
COUNT & --- & --- & 326.3 (95.9\%) & 18,586 (96.7\%) \\
\bottomrule
\end{tabular}

\end{table}

\begin{table}[htbp]
 \centering
 \caption{Peak GPU memory in GiB per executing GPU for each throughput measurement, as allocated (reserved) high-water marks of the PyTorch allocator. Training is sharded across eight H200s; inference runs on one H200. The learned n-gram baseline's GPU-resident table holds 2.16 GB.}
 \label{tab:throughput-memory}
 \setlength{\tabcolsep}{4pt}
 \begin{tabular}{@{}lrrrr@{}}
\toprule
Workload & Batch & Standard transformer & Learned n-gram & COUNT \\
\midrule
\multicolumn{5}{l}{\textit{GPU-resident n-gram storage}} \\
Training & 32 & 60.8 (61.7) & 63.4 (64.6) & --- \\
Prefill & 1 & 4.4 (6.1) & 6.4 (12.2) & --- \\
Prefill & 8 & 14.4 (20.1) & 16.4 (21.9) & --- \\
Decode & 1 & 4.0 (4.9) & 6.0 (6.9) & --- \\
Decode & 256 & 47.4 (110.8) & 49.4 (115.1) & --- \\
\midrule
\multicolumn{5}{l}{\textit{Host-resident n-gram storage}} \\
Training ($J=6$, 17.1 GB) & 32 & 60.8 (61.7) & --- & 61.0 (62.0) \\
Training ($J=1$, 17.1 GB) & 32 & 60.8 (61.7) & --- & 60.9 (62.0) \\
Training ($J=6$, 797.5 GB) & 32 & 60.8 (61.7) & --- & 61.0 (62.0) \\
Training ($J=1$, 797.5 GB) & 32 & 60.8 (61.7) & --- & 60.9 (62.0) \\
Prefill & 1 & 4.4 (6.1) & 4.4 (4.8) & 4.4 (6.1) \\
Prefill & 8 & 14.4 (20.1) & 14.4 (19.5) & 14.4 (18.9) \\
Decode & 1 & 4.0 (4.9) & 4.0 (4.7) & 4.0 (4.9) \\
Decode & 256 & 47.4 (110.8) & 47.4 (115.1) & 47.4 (105.4) \\
\bottomrule
\end{tabular}

\end{table}

\subsection{Table swap interpolation on loss}
\label{app:domain-swap-grid}

\begin{figure}[htbp]
 \centering
 \includegraphics[width=\linewidth]{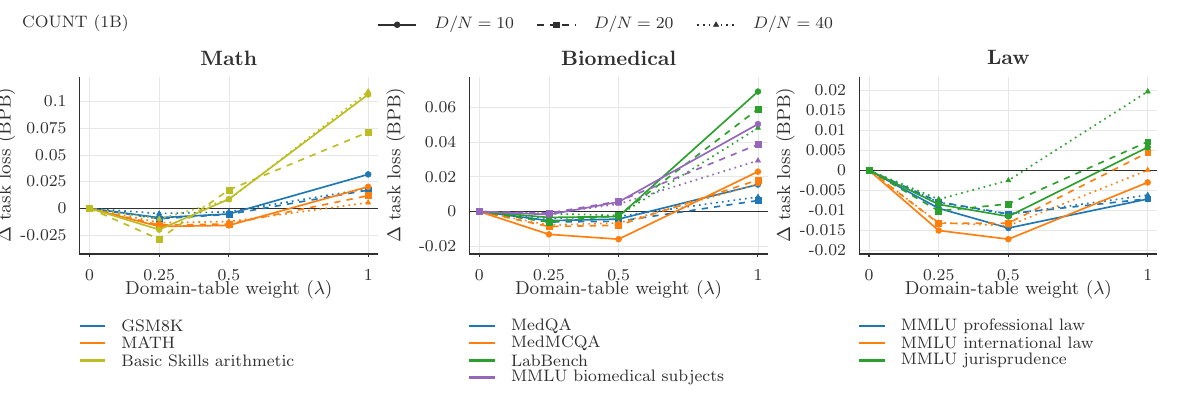}
 \caption{Full domain-table interpolation grid for 1B COUNT at $D/N=10,20,40$. Each metric is change relative to its own checkpoint's original-table score.}
 \label{fig:domain-swap-full-grid}
\end{figure}

Figure~\ref{fig:domain-swap-full-grid} shows the absolute task-loss changes at all four interpolation weights, averaging subjects within MATH, LabBench and the MMLU biomedical group.

\subsection{Inference-time $K$ and $J$ variation}
\label{app:inference-settings}
\label{sec:inference-settings}

\begin{figure}[htbp]
 \centering
 \includegraphics[width=\linewidth]{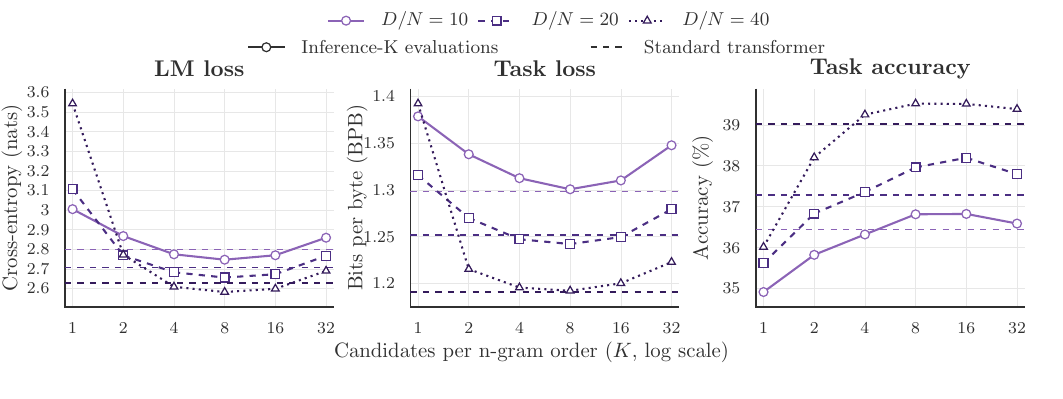}
 \caption{Inference-$K$ sweep for the 1B COUNT model with its original 797.487 GB table fixed.}
 \label{fig:inference-k-losses}
\end{figure}

\begin{figure}[htbp]
 \centering
 \includegraphics[width=\linewidth]{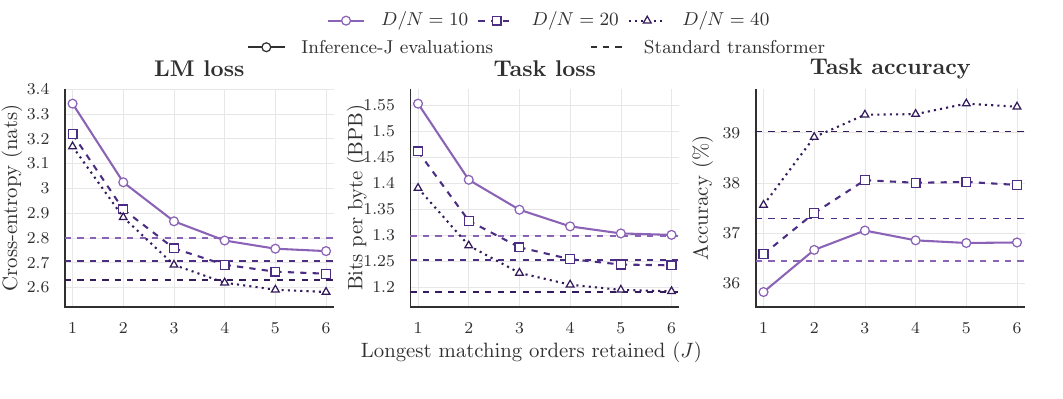}
 \caption{Inference-$J$ sweep for the 1B COUNT model with its original 797.487 GB table and $K=8$ fixed.}
 \label{fig:inference-j-losses}
\end{figure}

Starting from the trained 1B COUNT model, we test its sensitivity to two inference-time settings without retraining: the number of candidates kept per order, $K$, and the number of longest matching orders averaged, $J$ (Figures~\ref{fig:inference-k-losses} and \ref{fig:inference-j-losses}).
Both show some tolerance in LM loss and especially task accuracy. However, larger departures from the trained settings lose most of COUNT's improvement over the standard transformer.
A sensitivity sweep that pretrains each configuration may retain more of the improvement.

\subsection{Training with fewer n-gram orders}
\label{app:training-max-order}

\begin{figure}[htbp]
 \centering
 \includegraphics[width=\linewidth]{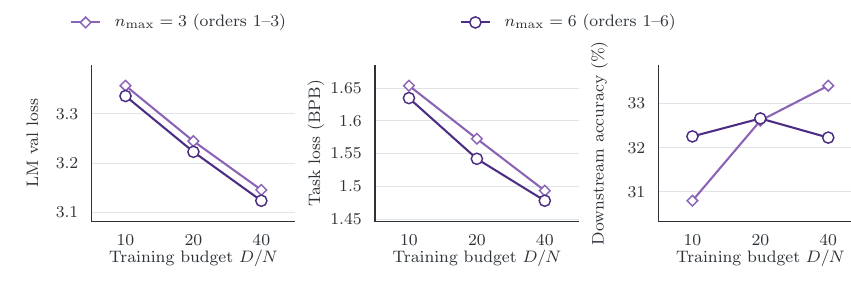}
 \caption{190M COUNT trained with $n_{\max}=3$ versus $n_{\max}=6$ maximum n-gram order with same 17.1 GB table, $K=8$, and embedding amplification.}
 \label{fig:max-order-190m}
\end{figure}

Figure~\ref{fig:max-order-190m} compares 190M COUNT trained with orders one through three against orders one through six, with the same table and $K$.
Six orders give lower LM and task loss at every budget, so we train with all six.

\subsection{Ablating embedding amplification}
\label{app:embedding-amplification}

\begin{figure}[htbp]
 \centering
 \includegraphics[width=\linewidth]{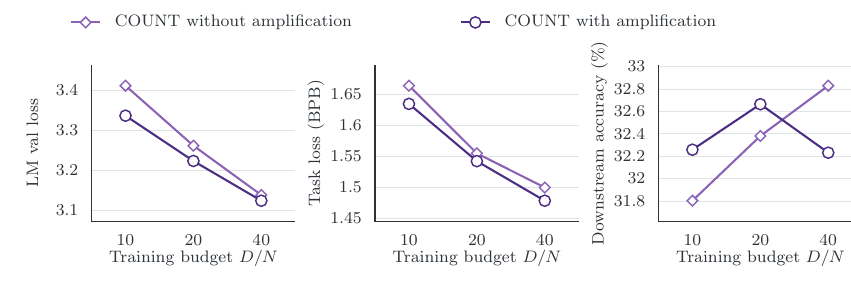}
 \caption{190M COUNT with and without embedding amplification (17.1 GB table, $K=8$, $J=6$).}
 \label{fig:embedding-amplification-count}
\end{figure}

\begin{figure}[htbp]
 \centering
 \includegraphics[width=\linewidth]{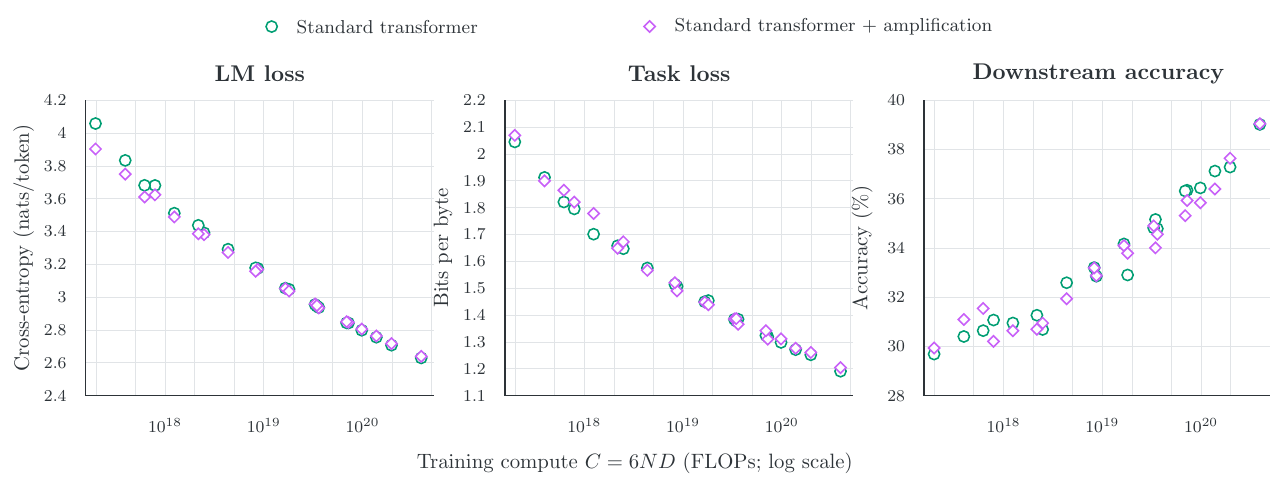}
 \caption{Embedding amplification across standard transformer scales at $D/N=10,20,40$.}
 \label{fig:vanilla-embedding-amplification}
\end{figure}

COUNT applies embedding amplification, $\sqrt{d}\,\operatorname{LayerNorm}(\cdot)$, to its combined input.
Figure~\ref{fig:embedding-amplification-count} ablates it within 190M COUNT: amplification lowers both LM and task loss at every budget.
Figure~\ref{fig:vanilla-embedding-amplification} adds amplification to the standard transformer itself across the ladder: it improves LM loss at the smallest scales, has mixed effects on downstream accuracy, and increases LM and task loss at 760M and 1B.
We therefore compare against the standard transformer without amplification.

\subsection{Ablating the COUNT embedding basis}
\label{app:count-basis}

\begin{figure}[htbp]
 \centering
 \includegraphics[width=\linewidth]{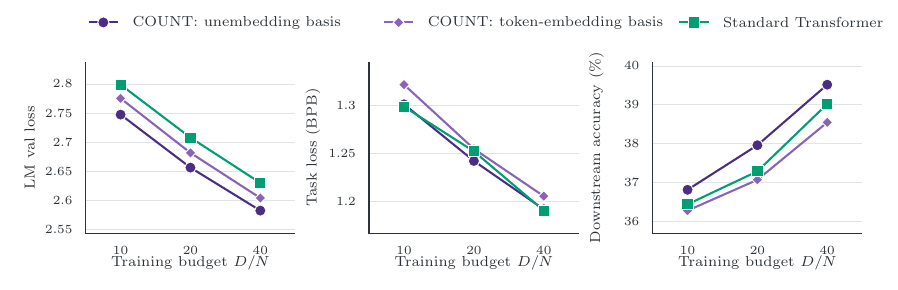}
 \caption{1B COUNT with an unembedding or token-embedding basis.}
 \label{fig:count-basis-1b}
\end{figure}

Figure~\ref{fig:count-basis-1b} compares the standard unembedding basis with reusing the input token embeddings as COUNT's basis at 1B.
The unembedding basis gives lower LM and task loss and higher accuracy at all three training budgets.
The token-embedding basis still lowers LM loss relative to the standard transformer, but has worse task loss and accuracy.

\end{document}